\documentclass[10pt,twocolumn,letterpaper]{article}

\usepackage{wacv}              

\usepackage{graphicx}
\usepackage{amsmath}
\usepackage{amssymb}
\usepackage{booktabs}

\newcommand{\NewDataset}[1]{MTReD}

\usepackage[pagebackref,breaklinks,colorlinks]{hyperref}

\usepackage[capitalize]{cleveref}
\crefname{section}{Sec.}{Secs.}
\Crefname{section}{Section}{Sections}
\Crefname{table}{Table}{Tables}
\crefname{table}{Tab.}{Tabs.}

\def\wacvPaperID{3} 
\def\confName{WACV}
\def\confYear{2025}

\begin{document}

\title{MTReD: 3D Reconstruction Dataset for Fly-over Videos of Maritime Domain}

\author{Rui Yi Yong$^{1,2}$, Samuel Picosson$^1$, Arnold Wiliem$^{1,3}$
{\tt\small  }
\\ 
$^1$Shield AI \{rui.yi, samuel.picosson, arnold.wiliem\}@shield.ai \\
$^2$Monash University, Melbourne, Australia \\
$^3$Queensland University of Technology, Queensland, Australia
}  


\maketitle

\begin{abstract}
This work tackles 3D scene reconstruction for a video fly-over perspective problem in the maritime domain, with a specific emphasis on geometrically and visually sound reconstructions. This will  allow for downstream tasks such as segmentation, navigation, and localization. To our knowledge, there is no dataset available in this domain. As such, we propose a novel maritime 3D scene reconstruction benchmarking dataset, named as MTReD (Maritime Three Dimensional Reconstruction Dataset). The MTReD comprises 19 fly-over videos curated from the Internet containing ships, islands, and coastlines. As the task is aimed towards geometrical consistency and visual completeness, the dataset uses two metrics: (1) Reprojection error; and (2) Perception based metrics. We find that existing perception based metrics, such as Learned Perceptual Image Patch Similarity (LPIPS), do not appropriately measure the completeness of a reconstructed image.
Thus, we propose a novel semantic similarity metric utilizing DINOv2 features coined DiFPS (DinoV2 Features Perception Similarity).
We perform initial evaluation on two baselines: (1) Structured from Motion (SfM) through Colmap; and (2) the recent state-of-the-art MASt3R model. We find that the reconstructed scenes by MASt3R have higher reprojection errors, but superior perception based metric scores.
To this end, some pre-processing methods are explored and we find a pre-processing method which improves both the reprojection error and perception based score. We envisage our proposed MTReD to stimulate further research in these directions. The dataset and all the code will be made available in this \url{https://github.com/RuiYiYong/MTReD}.

\end{abstract}

\section{Introduction}
\label{sec:intro}

\begin{figure}[t]
  \centering
   \includegraphics[width=1.0\linewidth]{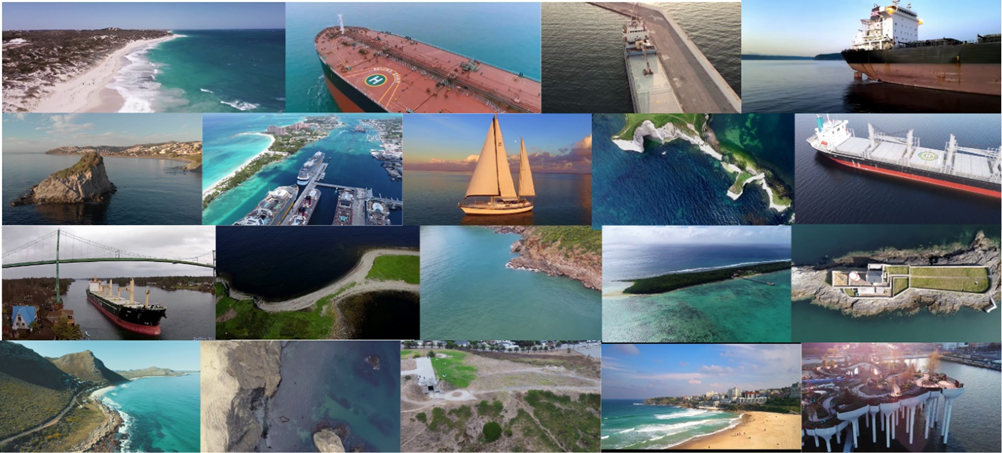}
   \caption{Some video frame examples from the proposed \NewDataset. dataset. The \NewDataset. contains multiple  Maritime settings such as coastal area, ships, islands. These settings pose unique challenges for the the general 3D reconstruction problem.}
   \label{fig:mtred_example}
\end{figure}

3D reconstruction is the process of recreating a scene or object’s geometry as a digital 3D model \cite{reviewOf3D}. In the context of computer vision, 3D reconstruction utilizes machine learning and algorithms to create models where input data is limited, often only to a set of input images. The application of 3D reconstruction spans many fields encompassing medicine \cite{tomographic}, land surveying \cite{land}, robotic navigation \cite{RN45}, and gaming \cite{FERDANI2020129}.

Despite steady progress in the field, to our knowledge, 3D scene reconstructions from an aerial perspectives has not been explored in the Maritime domain. Addressing this problem will pave the way for improvements to navigation, localization, and other pertinent downstream tasks with physical applications. For instance, increased reconstruction capabilities in coastal areas would allow for aerial drones to fly closer to land when performing search and rescue tasks. From a 3D reconstruction view, the Maritime domain houses a unique set of challenges ranging from dynamic sea regions, to the lack of feature points in sea and sky settings.

Hence, to stimulate research in this work, we propose a novel dataset, MTReD (Maritime Three Dimensional Reconstruction Dataset). 
MTReD is a video based scene reconstruction dataset which covers various maritime settings of ships, islands and coastlines. Captured from an aerial perspective, this dataset will be particularly useful in the context of unmanned aerial vehicles (UAV) and aerial based maritime searches. Figure~\ref{fig:mtred_example} shows some examples of the proposed \NewDataset w dataset.

With an emphasis on geometrically and visually sound reconstructions, our test evaluation protocol uses two main metrics: (1) Reprojection error\cite{7780814} and (2) Perception based metrics\cite{zhang2018unreasonableeffectivenessdeepfeatures}.
The average reprojection error is commonly used in the reconstruction field to evaluate the quality of reprojected images and estimated camera poses. While effective at evaluating the geometric consistency of these individual points and camera poses, it fails to consider scene completion.

The second metric uses deep neural network features to compute similarities between a reconstructed scene and original images. This similarity acts as a proxy to evaluate the completeness of a reconstruction. One of the metrics that fall within this category is LPIPS\cite{zhang2018unreasonableeffectivenessdeepfeatures}. Notwithstanding, we demonstrate in our experiments that the LPIPS metric does not faithfully reflect levels of scene completion. Specifically, two reconstructed images which have vastly different completeness levels, might be minimally affected in their LPIPS score. To this end, we propose a metric based on DINOv2\cite{oquab2024dinov2learningrobustvisual} features. Compared to LPIPS which is based on older model architectures such as VGG\cite{simonyan2015deepconvolutionalnetworkslargescale}, the DINOv2 architecture is more modern and it is trained using internet-scale training data.


Finally, we propose a method of pre-processing images to improve the reconstruction quality of recent state-of-the-art (SOTA) MASt3R\cite{leroy2024groundingimagematching3d} outputs, through leveraging a traditional Structured from Motion (SfM) implementation called Colmap\cite{7780814}. We provide a comparison with alternative pre-processing methods and use standard evaluation metrics alongside perception based metrics. This allows us to demonstrate the importance of these metrics in the context of end-to-end reconstruction pipelines, providing an intuitive yet effective metric to evaluate output scenes.\\

\noindent \textbf{Contributions} We list our contributions as follows.
\begin{enumerate}
\item A novel 3D reconstruction benchmarking dataset in the maritime domain - MTReD (Maritime Three-Dimension Reconstruction Dataset)
\item A novel perception based metric utilizing DINOv2 features~\cite{oquab2024dinov2learningrobustvisual} coined DiFPS (DinoV2 Features Perception Similarity). When compared to LPIPS, it demonstrates superior ability in evaluating scene completion while retaining equal sensitivity to scene accuracy   
\item A study on image pre-processing methods along with identification of a recommended pipeline to optimize geometric consistency and scene completion in MASt3R reconstructions
\end{enumerate}

We continue our paper as follows. Section~\ref{sec:related work} will first focus on related works within the reconstruction field, before Section~\ref{sec:Dataset} focuses on the background of the proposed MTReD along with the pre-processing methods and evaluation metrics to be used. Experiment setups will then be highlighted in Section~\ref{sec:Experiments}. Lastly, results and their implications will be discussed in ~\ref{sec:Results}.

\section{Related Work}
\label{sec:related work}

We first discuss works related to image based datasets, initially looking at outdoor datasets specific to 3D reconstruction, before focusing on Maritime Related Datasets. In both cases we highlight their relevant use cases and issues which result in the need for \NewDataset w. The field of 3D reconstruction is then explored, before concluding with a discussion of metrics along with motivations for a proposed novel perception metric.\\

\noindent
\textbf{Outdoor Aerial 3D Reconstruction Datasets -} There are several datasets designed for 3D reconstruction which contain image sequences of outdoor scenes. For instance, Blended MVS\cite{yao2020blendedmvslargescaledatasetgeneralized}, DTU\cite{jensen2014large}, and Tanks and Temples\cite{Knapitsch2017} are common datasets used for benchmarks in the reconstruction field. Notwithstanding, scenes from these datasets are not always captured from aerial perspectives. The recent Outdoor Multi-modal dataset (OMMO)\cite{lu2023largescaleoutdoormultimodaldataset} provides good coverage of complex outdoor scenes from aerial viewpoints, however this dataset does not have a specific focus on maritime contexts. As such, while outdoor aerial 3D reconstruction datasets do exist, there is no dataset specific to Maritime settings.  \\

\noindent
\textbf{Maritime-Related Datasets -} The Seagull \cite{seagull} and  SeaDronesSee\cite{varga2021seadronesseemaritimebenchmarkdetecting} datasets both provide outdoor, aerial based image sequences of boats and objects floating in the sea. However, these datasets do not provide adequate scene coverage for effective 3D reconstruction. Moreover, each dataset was captured professionally and will not be effective in testing the robustness of reconstruction methods from different scene makeups and capture equipment. Lastly, these datasets were not captured with 3D reconstruction in mind, instead focusing on object tracking and classification. As a result, scenes are often overly simplistic and lack sufficient features to be used in 3D reconstruction contexts. \\

\noindent
\textbf{3D reconstruction in the general vision domain -} Considering SOTA 3D reconstruction pipelines, the current standard to obtain point clouds and estimated camera poses is SfM and its variants, being explicitly specified in the pipeline of both NeRF \cite{mildenhall2020nerfrepresentingscenesneural} and Gaussian Splatting \cite{kerbl20233dgaussiansplattingrealtime} papers. In particular, incremental SfM has found large success and is implemented through an open-source library, Colmap.\\ 
\\Considering recent advancements, the 2024 ECCV work on Dense and Unconstrained Stereo 3D Reconstruction (DUSt3R) \cite{wang2024dust3rgeometric3dvision} presents a fresh approach. DUSt3R reduces the reconstruction problem to a regression of point maps using a standard transformer network. These point maps are a one-to-one mapping between image pixels and points within a 3D scene. Matching And Stereo 3D Reconstruction (MASt3R) is a network built on DUSt3R more specialized for camera pose estimation and reconstruction tasks \cite{leroy2024groundingimagematching3d}. An additional local feature head allows the network to jointly optimize image matching with reconstructions. MASt3R claims SOTA performance on standard reconstruction benchmarks, including lower Visual Correspondence Reconstruction Errors (VCRE) and higher percentages in estimated camera pose accuracies.  
Despite these advancements, the models above have yet to be tested within the context of outdoor aerial maritime scenes. Hence, this work will use the proposed MTReD to provide an initial study on these baselines.\\

\noindent
\textbf{3D reconstruction evaluation metrics - } When ground truth data is unavailable, the most common error metric used in the 3D reconstruction field is reprojection error\cite{7780814}. The Reprojection error provides a reliable indication on the geometric accuracy of reconstructions, however it fails to provide any information on visual soundness and scene completion. To address this, one potential metric to use is the Learned Perceptual Patch Similarity  (LPIPS) score\cite{zhang2018unreasonableeffectivenessdeepfeatures}. LPIPS aims to mimic human perception in its evaluation of how similar two images are. While it has been used in previous works such as NeRF\cite{mildenhall2020nerfrepresentingscenesneural}, scene completion was not necessary to consider as NeRF outputs values for every pixel in an image. Moreover, LPIPS lacks a semantic understanding of scenes, being more sensitive to color and texture. This means that changes in illumination from different capture angles will be penalized more harshly with LPIPS. \cite{fu2023dreamsimlearningnewdimensions}. Furthermore, although the LPIPS metric uses pre-trained backbones such as VGG\cite{simonyan2015deepconvolutionalnetworkslargescale}, it requires training on its heads.
Considering this, we propose that a DINOv2\cite{oquab2024dinov2learningrobustvisual} based perception score be used, and argue that as DINOv2 is trained using internet-scale levels of data, additional training data is not required.
The semantic awareness of DINOv2 is also especially useful in creating a global coherence to evaluate whether scenes make sense wholistically. In our experiments, we demonstrate that our proposed metric is able to more accurately follow image perception by jointly evaluating visual soundness and scene completion.

\section{MTReD}
\label{sec:Dataset}

This section will outline the proposed MTReD along with the 
metrics used for benchmarking and pre-processing methods.

\NewDataset w consists of frames extracted from 19 open-source YouTube videos, containing three maritime scenes: (1) ships (seven videos); (2) islands (five videos); and (3) coastlines (seven videos). Note that within one video, multiple scenes may exist (e.g., ship and coastline scenes). In this setting, we categorize the video by its predominant scene. Each scene type has its own set of unique challenges. Ship scenes have a lack of feature points, especially when in the middle of the ocean. Meanwhile, islands and coastlines are richer in feature points, but the similarity of these points also means they are more prone to artifacts during 3D reconstruction.
The use of open-source data is intentional and allows for data to be more akin to amateur captured videos. The motivation for this is to create a dataset which better represents the wealth of online data. Following this line of reasoning, scenes might also contain small dynamic movement in background regions. However, the MTReD dataset does not cover scenarios where there are large dynamics (e.g., fast moving objects in the air or on the sea water).
Examples of the MTReD can be seen in Figure~\ref{fig:mtred_example}.
For each video we first discard any irrelevant frames (e.g., fade in/out transitions and title frames). The proposed MTReD is then created by saving every $N$ frame from each video. More details about each video and the selected frames are available in the supplementary materials.

\subsection{Evaluation metrics}
MTReD uses two main evaluation metrics: (1) the reprojection error~\cite{7780814}; and perception based metrics. 
The proposed MTReD does not come with ground truth data. Thus, both evaluation metrics exploit the continuity of input frames to ensure that reconstructed objects are part of the same scene. In other words, once a 3D scene has been reconstructed, one can pick any video frame used in the 3D reconstruction process and perform reprojection of the 3D scene into a 2D image of the selected video frame.\\


\noindent
\textbf{Reprojection Error } - This is a common metric used in the reconstruction field when there is a lack of ground truth data \cite{7780814}. The reprojection error evaluates geometric consistency through point tracking, and distance computations of projected 3D points against their original positions in input images. 
The reprojection error is defined as follows
\begin{equation}
\operatorname{Reprojection Error} =\|\textbf{p} - \hat{\textbf{p}} \|^2 , 
\end{equation}
where $\textbf{p}$ represents the pixel coordinates of a point in the original image, and $\hat{\textbf{p}}$ represents the pixel coordinates of the associated and optimized 3D point, projected onto an estimated original image plane. Essentially, this computes the $\ell_2$ norm for projected point coordinates against their original positions.
We argue that while the reprojection error metric is effective at indicating the accuracy of camera pose estimations, it does not evaluate scene completion. Additionally, it also fails to penalize scores from unwanted artifacts, a problem which will be further highlighted in the discussion. This leads to the need of perception based metrics.\\

\noindent
\textbf{Perception based metrics } - These metrics aim to measure the similarity between two images~\cite{perceptualLoss2016CVPR,zhang2018unreasonableeffectivenessdeepfeatures} along the same vein of human perception. In MTReD, perception based metrics are used to measure reconstruction accuracy along with scene completion (\textit{i.e.,} a comparison of reprojected images with their reference images).
A common perception metric is the Learned Perceptual Image Patch Similarity (LPIPS) which uses pre-trained deep neural network backbones such as VGG\cite{simonyan2015deepconvolutionalnetworkslargescale} and AlexNet \cite{NIPS2012_c399862d}. LPIPS also trains a calibrator layer on the top of the backbones. While multiple training protocols exist, for our chosen implementation, we use the linear variation where only the added head is trained. We refer readers to the LPIPS paper in~\cite{zhang2018unreasonableeffectivenessdeepfeatures} for further details. 
Notwithstanding, LPIPS uses dated models with subpar amounts of training data compared to recent standards. Hence, we propose the use of more modern architectures with higher levels of semantic understanding, such as Transformer networks\cite{dosovitskiy2021imageworth16x16words} trained on internet-scale data. Specifically, we opt to use DINOv2 and thus we name this as the DiFPS metric. 
Due to strong features exhibited by DINOv2 we further simplify calculations by excluding a calibration layer. This means, once the DINOv2 features are extracted we simply compute a cosine similarity value as our metric as follows
\begin{equation}
\operatorname{DiFPS} ( \textbf{I}_1, \textbf{I}_2 ) = \frac{\textbf{X}_1 \cdot \textbf{X}_2}{|| \textbf{X}_1 || || \textbf{X}_2 ||} , 
\end{equation}
where $\textbf{I}_1$ and $\textbf{I}_2$ are the input images; $\textbf{X}_1$ and $\textbf{X}_2$ are the corresponding DINOv2 features; and $\| \cdot \|$ is the $\ell_2$ norm.\\

\noindent
\textbf{Additional metrics - } The proposed MTReD also uses several other metrics such as: the average image throughput, and point count per image. The average image throughput indicates the number of video frames used in the reconstruction. The point count per image refers to the number of point clouds generated per image. Both metrics are supplementary to the main metrics and aid in finer analysis.\\


\subsection{Pre-processing methods}
As the MTReD consists of video-extracted images, we often have a surplus of input frames and do not require every frame for successful reconstruction. Furthermore, inter-frame variations from illumination, exposure, and other factors, may adversely affect the reconstruction result. Here we study several pre-processing strategies and evaluate their efficacy using the dataset evaluation metrics. These strategies are as follows: (1) Colmap based filtering; (2) Contrast based normalization; and (3) Background based normalization.\\

\noindent
\textbf{Colmap based filtering - } This pre-processing method act as a simple filtering mechanism where we only select video frames which Colmap selects during reconstruction~\cite{7780814}. Colmap has stringent requirements in selecting images for reconstruction\cite{wang2024dust3rgeometric3dvision}. Thus, we assume that the selected images should be of high quality for reconstruction purposes.\\



\noindent
\textbf{Contrast based normalization - }
We observe that there are contrast variations between video frames due to differences in illumination intensity. Specifically, models might struggle to find matches in scenarios where low contrast makes feature points more uniform in their pixel value distribution. To this end, we increase the contrast by a fixed value to improve feature detection using contrast based normalization. \\
 

\begin{figure}
  \centering
   \includegraphics[width=0.95\linewidth]{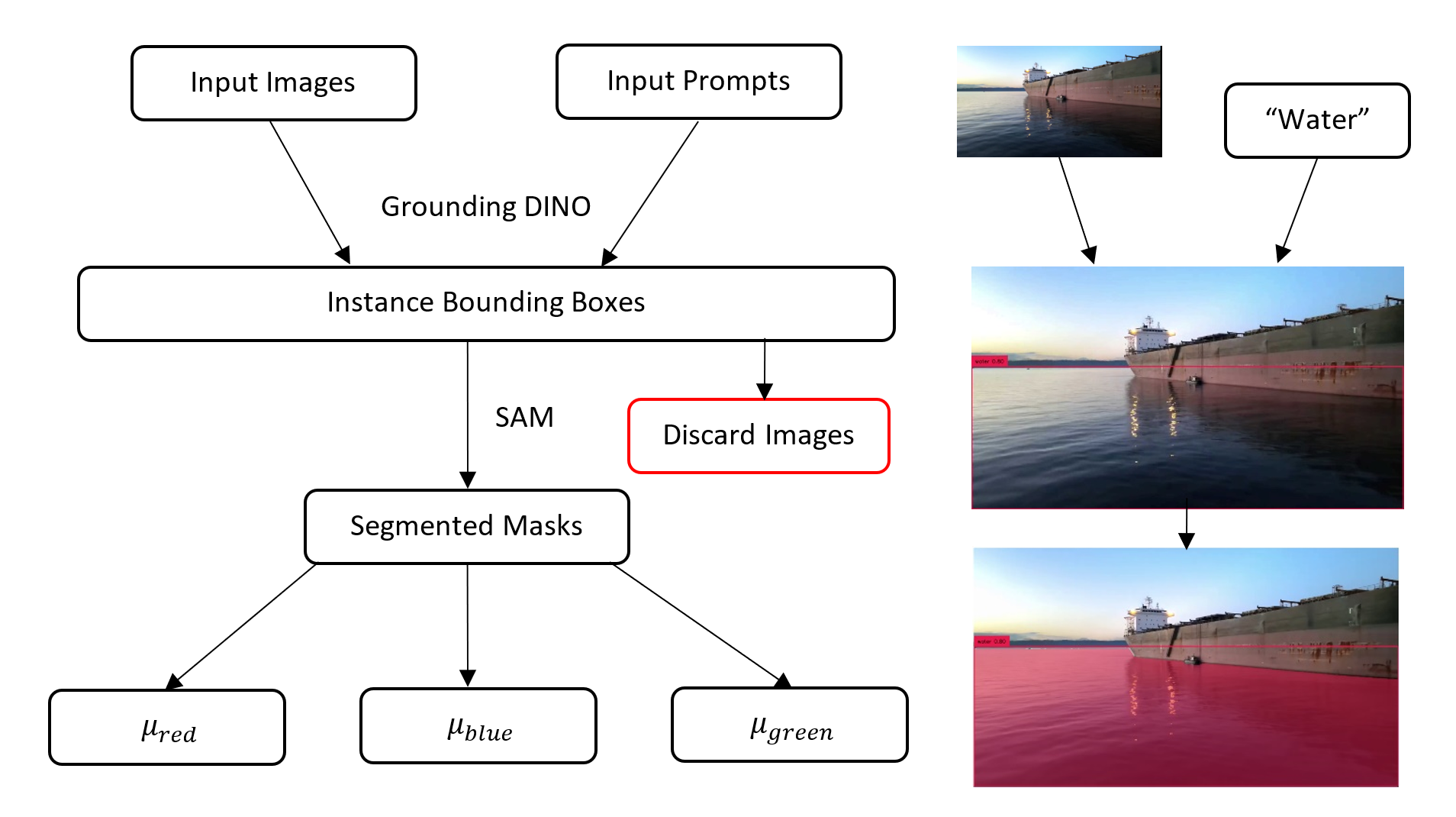}
   \caption{A Grounding DINO to SAM pipeline used for background segmentation along with image examples. The collected information on mean and standard deviation for each color channel is also highlighted. This information is then  used for image processing.}
   \label{fig:backgroundSeg}
\end{figure}

\begin{table*}
  \centering
  {\small{
  \begin{tabular}{p{3cm}|p{2.5cm}|p{2.5cm}|p{2.5cm}|p{2cm}|p{2cm}}
     \toprule
     Dataset Type & Image Throughput (in \%) $\uparrow$ & Reprojection Error $\downarrow$ & Point Count per Image $\uparrow$ & LPIPS $\downarrow$  & DiFPS $\uparrow$ \\ \hline
    Colmap & 65.4\% & \textbf{0.859} & 324 & 0.485 & 0.560\\ \hline
    MASt3R - No pre-processing & \textbf{100.0\%} & 0.914 & \textbf{21050} & 0.435 & 0.779 \\ \hline
    MASt3R - Colmap Based Filtering (proposed) & 65.4\% & 0.874 & 16608 & \textbf{0.410} & \textbf {0.803} \\ \hline
  \end{tabular}
  }}
  \caption{A comparison of Colmap against MASt3R reconstructions with unprocessed images and a proposed Colmap Based Filtering method. The best results for each column are bolded.}
  \label{tab:tableShort}
\end{table*}

\noindent
\textbf{Background based normalization - }
The pipeline for this method is shown in Figure~\ref{fig:backgroundSeg}.
We exploit common background regions in Maritime settings, such as the sky and sea, as an alternative approach to tackle inter-frame variations. Specifically, we assume that the pixel statistics of these regions should remain consistent across images. 
To enforce this, we first identify sky or sea regions in each image frame using Grounding DINO~\cite{liu2023grounding}. If these regions are not found, the image will be discarded. The bounding boxes of the detected regions are then fed into SAM~\cite{kirillov2023segany} to generate more accurate region masks. The mean pixel value for each mask is then computed and stored. 
Finally, the normalization is done as follows
\begin{equation}
    \hat{I_i}(x,y,c) = \frac{\mu^c}{ \mu^c_i } I_i(x,y,c) , 
\end{equation}
where $I_i(x,y,c)$ is the the pixel value at location $(x,y)$ with color channel $c$ of image $i$; $\hat{I_i}(x,y,c)$ is the normalized pixel value of image $i$; ${\mu^c}$ then represents the region mean pixel value of channel $c$ across all input video frames; and ${\mu^c_i}$ represents the region mean pixel value for the $i$-th image.
There are two variants for this method: (1) White Balance - Sky; and (2) White Balance - Water. The White Balance - Sky uses sky regions, and the White Balance - Water uses water/sea regions. 

\section{Experiment Setup}
\label{sec:Experiments}

We divide our experiments into three sections: (1) results on baselines and a comparison of Colmap and MASt3R; (2) exploration of pre-processing methods; and (3) validating the proposed DiFPS metric.\\

\noindent
\textbf{Colmap~\cite{7780814} - } is a general-purpose SfM framework. It uses the most prevalent SfM strategy for 3D reconstruction from unordered image collections. We use Colmap from its standard open-source implementation available at~\url{https://github.com/colmap/colmap} and utilize the default parameters for performing reconstruction.\\

\noindent
\textbf{MASt3R~\cite{leroy2024groundingimagematching3d} - } is the recent SOTA method utilizing a deep neural network architecture to perform various 3D vision tasks. As mentioned before, MASt3R is based on DUSt3r with an additional dense local feature head which allows significantly more accurate reconstructions with lesser computational complexities. We use the official MASt3R implementation available at~\url{https://github.com/naver/mast3r}. Again, default parameters are also used during the evaluation. \\

\noindent
\textbf{Evaluation protocol - } Given image frames extracted from a video, we first obtain Colmap and MASt3R reconstructions for the unprocessed data. We then use each processing method and feed outputs into MASt3R. Point clouds and estimated camera poses are then extracted for every reconstruction. Each estimated camera pose correlates to an input frame, and we compute the evaluation metrics for each pair. In practice, the reprojection error is provided by both Colmap and MASt3R. Thus, we use the reprojection error values from their respective implementations. The average metric values across each dataset are then reported. \\


\noindent
\textbf{Perception based metrics - } We use the official implementation for DINOv2 available in~\url{https://github.com/facebookresearch/dinov2}. The pytorch lighting implementation (i.e., the torchmetrics.image.lpip) is then used for LPIPS. Once a 3D scene has been reconstructed, we first reproject each 3D point cloud into the estimated camera poses of every participating input image. Note that each projected point is circularly expanded by factor of 3 pixels. Distance to the camera is also accounted for to ensure that points in closer proximity take priority in the case of point overlap. Both perception metrics are then computed by comparing reprojected images against their corresponding original input images.  \\


\noindent
\textbf{Pre-processing methods - } Using the evaluation metrics applied on MTReD, we compare all four variants of pre-processing methods : (1) Colmap based filtering; (2) Contrast based normalization; (3) White balance - Sky; and (4) White balance - Water. Background based pre-processing methods use the default parameters available in Grounding DINO and SAM. Contrast based pre-processing methods use the OpenCV function called $\operatorname{addWeighted}$, with an alpha value of 1.8.




\section{Results}
\label{sec:Results}

\begin{figure*}
  \centering
   \includegraphics[width=0.95\linewidth]{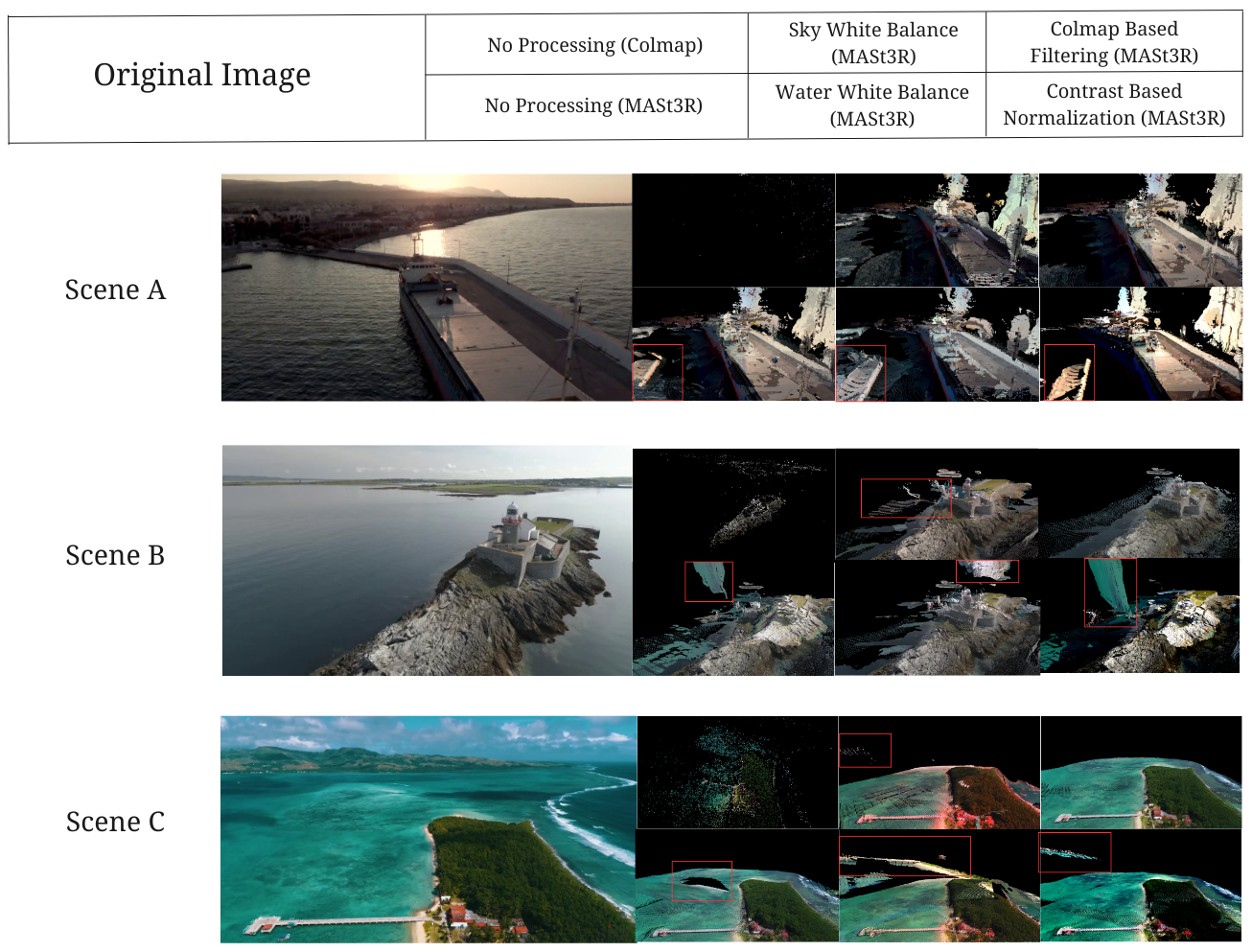}
   \caption{A comparison of reprojected images against original input images. The first table maps the layout of each reconstruction type in the subsequent images. With the exception of the Colmap reprojections, all other reprojections use a MASt3R reconstruction. Red rectangles are used to highlight major reconstruction artifacts in each reprojected image.}
   \label{fig:comparison}
\end{figure*}

We first show the comparisons between Colmap and MASt3R on our proposed dataset, MTReD. Then we show comparisons for the four variants of pre-processing methods along with the unprocessed dataset. Finally, we evaluate our proposed DiFPS metric against the existing LPIPS metric. The Colmap and MASt3R comparisons results serve as the baselines for future work. 

The results in Table~\ref{tab:tableShort} compare Colmap with two variants of MASt3R, one without pre-processing, and the other using the proposed Colmap Based Filtering method.
While Colmap is largely used as a pre-processing step in modern reconstruction pipelines, it acts as the primary stage of geometric interpolations within these pipelines, estimating camera poses along with common points in the 3D space. This makes it a suitable baseline to consider when approaching the reconstruction problem from a standpoint of geometric consistency and visual soundness.

From the table, Colmap achieves a lower average reprojection error. This indicates that the points found by Colmap are more geometrically consistent with input image pixels. Notwithstanding, the lack of sufficient points leads to a lower final image quality, indicated by both perception based metrics. As MTReD is not strictly static in scene makeup, Colmap is expected to be more inflexible to these variations being an algorithm based method.

It is also important to note that the range of reprojection error values is not bounded between 0 and 1. Thus, the difference in reprojection errors between MASt3R and Colmap is not as significant.
Furthermore, MASt3R's superior perception scores indicate that despite its slightly higher reprojection errors, it still produces more semantically complete reconstructions. Visual inspection of the images also indicates that MASt3R is able to provide an acceptable level of geometric consistency while greatly increasing the level of scene completion. A comparison of reprojected images against their input images is also provided in Figure~\ref{fig:comparison}. 

\subsection{Baseline evaluation}
\begin{table*}[t]
  \centering
  {\small{
  \begin{tabular}{p{3cm}|p{2.5cm}|p{2.5cm}|p{2.5cm}|p{2cm}|p{2cm}}
     \toprule
     Dataset Type & Image Throughput (in \%) $\uparrow$  & Reprojection Error $\downarrow$ & Point Count per Image $\uparrow$ & LPIPS $\downarrow$ & DiFPS $\uparrow$ \\ \hline
    No pre-processing & \textbf{100.0\%} & 0.914 & 21050 & 0.435 & 0.779\\ \hline
    Colmap Based Filtering & 65.4\% & 0.874 & 16608 & \textbf{0.410} &\textbf{0.803}\\ \hline
    Contrast & \textbf{100.0\%} & 0.989 & 20111 & 0.455 &0.773\\ \hline
    White Balance - Sky & 61.4\% & 0.886 & 22487 & 0.426&0.764\\ \hline
    White Balance - Water & 69.9\% & \textbf{0.872} & \textbf{23172} & 0.426 &0.766\\ \hline
  \end{tabular}
  }}
  \caption{A comparison of pre-processing methods using a MASt3R reconstruction pipeline. Standard metrics are provided along with the last 2 columns providing perception based metrics. The best results for each column are highlighted in bold.}
  \label{tab:tableLong}
\end{table*}

\subsection{Pre-processing method evaluation}

Table~\ref{tab:tableLong} contains the evaluation results of MASt3R with different pre-processing methods. 
The first row represents the variant of MASt3R without any pre-processing, with subsequent rows representing each of the aforementioned pre-processing methods. Image throughput indicates the number of images used in the reconstruction process. As expected, MASt3R always has a 100\% throughput, and reduction in these values are the result of employing pre-processing filtering techniques. 

Both perception based scores had their best results when a Colmap Based Filtering method was used. The intuition behind this result is that images which Colmap does not select are likely to be detrimental for the reconstruction process. This could be due to a variety of reasons including difficulties in feature point matching or a lack of overlapping regions. As a result, forcing these images on a reconstruction might degrade its quality in the form of unwanted artifacts. This is apparent from scene (A) in Figure~\ref{fig:comparison}, where an extra jetty artifact can be seen in the red boxes. Using the proposed Colmap Based Filtering method, we see this artifact disappears. Further, when we consider scene (B) in Figure~\ref{fig:comparison}, apart from the Colmap Based Filtering method, every other MASt3R variant has artifacts present in them. Hence, the proposed Colmap Based Filtering method appears the most geometrically consistent across all three scenes in the figure. This is also reflected by superior scores when considering the perception based metrics. 

We also confirm that a major difficulty in the reconstruction of maritime scenes lies in the background sea and sky regions. The artifacts from reprojected images are all either part of or made up from these regions. This indicates that the textureless nature of backgrounds is a real challenge which can lead to unintended point matching and the creation of artifacts. Lastly, scene (C) in Figure ~\ref{fig:comparison} shows that as long as some level of texture exists, MASt3R is still able to reconstruct these regions to some extent. 

Despite the improved result from the proposed Colmap Based Filtering method, one key drawback is the reduction in the number of input images which potentially contain important views. However, considering the context of physical applications, geometric consistency takes priority over scene coverage as this will at least provide a region for which the reconstructed model will be accurate and useful. 

\subsection{Perception based metric evaluation}
Looking back at the results in Table~\ref{tab:tableShort}, we compare the efficacy between LPIPS and DiFPS metrics in measuring the scene completeness.
Comparing between the two MASt3R variants, the improvements of both metrics are similar (i.e., 0.025 vs 0.024). However, when we compare Colmap with MASt3R, we see that significant improvement only occurs for the DiFPS metric. Looking at Figure ~\ref{fig:comparison}, we can see that there is actually a significant difference in the level of scene completion between these methods, and a perception based score should reflect this. In this aspect, we argue that the DiFPS metric is superior, having a similar sensitivity to LPIPS in terms of artifact handling, while concurrently providing more sensitivity to scene completion. 

We also highlight a major shortcoming of the reprojection error metric by considering unwanted artifacts. Looking at Figure~\ref{fig:comparison}, some major artifacts are highlighted within the red rectangles. However, these artifacts will have no impact on the reprojection error as it only calculates errors for pixels which exist in a corresponding input frame. Hence, the reprojection error fails to account for false positives and can be misleading in the context of major artifacts. 

Various pre-processing methods were successful in increasing the number of points per image, however it is apparent that this does not guarantee more complete reconstructions. In fact the Colmap Based Filtering method had the best DiFPS score while having the lowest point count per image. This can be explained by the reduction in unwanted artifacts leading to an elimination of unwanted and inaccurate points. These findings are further confirmed by the White Balance - Water method having the lowest reprojection error with the highest point count, yet producing one of the worst DiFPS scores. Hence, this highlights the importance of more wholistic evaluation techniques such as the perception metrics.

\section{Conclusion}
\label{sec:formatting}

In this paper, we propose a novel dataset, called MTReD which focuses on 3D reconstruction in Maritime scenes captured from aerial perspectives. As the target downstream tasks for this dataset are segmentation, navigation and localization, we use geometric consistency and scene completion as key determinants for success. 
We study two baselines: (1) Colmap, the traditional Structured-from-Motion (SfM) method; and (2) MASt3R, the recent SOTA method for 3D vision tasks. From our evaluation in MTReD, we find that although Colmap has excellent geometric consistency, measured by the reprojection error, it has a poor scene completeness, measured using perception based metrics. Colmap generates point clouds which are sparsely distributed in a 3D scene. In contrast, MASt3R provides detailed, explicit volumetric representation of scenes with competitive reprojection errors. While this makes MASt3R highly suited for downstream physical applications, future research should also consider using MASt3R as a pre-processing step for NeRF and Gaussian Splatting models. This will allow for a 100\% throughput of input images, while MASt3R's dense point-cloud could also improves initialization capabilities. 
In our work, we propose a novel perception based metric for measuring scene completeness. Specifically, our observations show that the existing LPIPS metric fails to decline proportionally when there are drops in scene completeness.
On the other hand, the proposed DiFPS metric shows much closer correlation with the quality of scene completeness. 
Unlike LPIPS, the DiFPS metric is based on more modern transformer architectures which are trained using internet-scale data.
Finally, our work also explores several pre-processing methods which are aimed towards improving 3D reconstruction quality by filtering out or normalizing variations in image statistics across input video frames. 
From this study we are able to improve the MASt3R reconstruction quality (both reprojection error and scene completeness) via utilizing images selected by Colmap.
While this approach may risk losing important views, this trade-off favors high geometric consistency over full scene coverage which is important in physical applications that focus on high levels of visual fidelity and local consistency.
We hope the proposed MTReD  and initial studies reported in this paper will pave the way for better 3D reconstruction algorithms in this domain.

{\small
\bibliographystyle{ieeetr} 

\begin{thebibliography}{10}

\bibitem{reviewOf3D}
A.~Moncef and A.~K. M'hamed, ``A review on 3d reconstruction techniques from 2d images,'' pp.~510--522, 02 2020.

\bibitem{tomographic}
U.~Khan, A.~Yasin, M.~Abid, I.~Shafi, and S.~Khan, ``A methodological review of 3d reconstruction techniques in tomographic imaging,'' vol.~42, p.~190, 09 2018.

\bibitem{land}
S.~Grunwald and P.~Barak, ``3d geographic reconstruction and visualization techniques applied to land resource management,'' vol.~7, pp.~231--241, 03 2003.

\bibitem{RN45}
Z.~Zhu, A.~Su, H.~Liu, Y.~Shang, and Q.~Yu, ``Vision navigation for aircrafts based on 3d reconstruction from real-time image sequences,'' {\em Science China Technological Sciences}, vol.~58, no.~7, pp.~1196--1208, 2015.

\bibitem{FERDANI2020129}
D.~Ferdani, B.~Fanini, M.~C. Piccioli, F.~Carboni, and P.~Vigliarolo, ``3d reconstruction and validation of historical background for immersive vr applications and games: The case study of the forum of augustus in rome,'' {\em Journal of Cultural Heritage}, vol.~43, pp.~129--143, 2020.

\bibitem{7780814}
J.~L. Schönberger and J.-M. Frahm, ``Structure-from-motion revisited,'' in {\em CVPR}, 2016.

\bibitem{zhang2018unreasonableeffectivenessdeepfeatures}
R.~Zhang, P.~Isola, A.~A. Efros, E.~Shechtman, and O.~Wang, ``The unreasonable effectiveness of deep features as a perceptual metric,'' 2018.

\bibitem{oquab2024dinov2learningrobustvisual}
M.~Oquab, T.~Darcet, T.~Moutakanni, H.~Vo, M.~Szafraniec, V.~Khalidov, P.~Fernandez, D.~Haziza, F.~Massa, A.~El-Nouby, M.~Assran, N.~Ballas, W.~Galuba, R.~Howes, P.-Y. Huang, S.-W. Li, I.~Misra, M.~Rabbat, V.~Sharma, G.~Synnaeve, H.~Xu, H.~Jegou, J.~Mairal, P.~Labatut, A.~Joulin, and P.~Bojanowski, ``Dinov2: Learning robust visual features without supervision,'' 2024.

\bibitem{simonyan2015deepconvolutionalnetworkslargescale}
K.~Simonyan and A.~Zisserman, ``Very deep convolutional networks for large-scale image recognition,'' in {\em ICLR}, 2015.

\bibitem{leroy2024groundingimagematching3d}
V.~Leroy, Y.~Cabon, and J.~Revaud, ``Grounding image matching in 3d with mast3r,'' in {\em ICCV}, 2024.

\bibitem{yao2020blendedmvslargescaledatasetgeneralized}
Y.~Yao, Z.~Luo, S.~Li, J.~Zhang, Y.~Ren, L.~Zhou, T.~Fang, and L.~Quan, ``Blendedmvs: A large-scale dataset for generalized multi-view stereo networks,'' 2020.

\bibitem{jensen2014large}
R.~Jensen, A.~Dahl, G.~Vogiatzis, E.~Tola, and H.~Aan{\ae}s, ``Large scale multi-view stereopsis evaluation,'' in {\em 2014 IEEE Conference on Computer Vision and Pattern Recognition}, pp.~406--413, IEEE, 2014.

\bibitem{Knapitsch2017}
A.~Knapitsch, J.~Park, Q.-Y. Zhou, and V.~Koltun, ``Tanks and temples: Benchmarking large-scale scene reconstruction,'' vol.~36, 2017.

\bibitem{lu2023largescaleoutdoormultimodaldataset}
C.~Lu, F.~Yin, X.~Chen, T.~Chen, G.~YU, and J.~Fan, ``A large-scale outdoor multi-modal dataset and benchmark for novel view synthesis and implicit scene reconstruction,'' 2023.

\bibitem{seagull}
M.~Marques, P.~Dias, N.~Pessanha~Santos, V.~Lobo, R.~Batista, D.~Salgueiro, A.~P. Aguiar, M.~Costa, J.~Estrela~da Silva, A.~Ferreira, J.~Sousa, M.~Nunes, E.~Pereira, J.~Morgado, R.~Ribeiro, J.~Marques, A.~Bernardino, M.~Grine, and M.~Taiana, ``Unmanned aircraft systems in maritime operations: Challenges addressed in the scope of the seagull project,'' pp.~1--6, 05 2015.

\bibitem{varga2021seadronesseemaritimebenchmarkdetecting}
L.~A. Varga, B.~Kiefer, M.~Messmer, and A.~Zell, ``Seadronessee: A maritime benchmark for detecting humans in open water,'' 2021.

\bibitem{mildenhall2020nerfrepresentingscenesneural}
B.~Mildenhall, P.~P. Srinivasan, M.~Tancik, J.~T. Barron, R.~Ramamoorthi, and R.~Ng, ``Nerf: Representing scenes as neural radiance fields for view synthesis,'' 2020.

\bibitem{kerbl20233dgaussiansplattingrealtime}
B.~Kerbl, G.~Kopanas, T.~Leimkühler, and G.~Drettakis, ``3d gaussian splatting for real-time radiance field rendering,'' 2023.

\bibitem{wang2024dust3rgeometric3dvision}
S.~Wang, V.~Leroy, Y.~Cabon, B.~Chidlovskii, and J.~Revaud, ``Dust3r: Geometric 3d vision made easy,'' 2024.

\bibitem{fu2023dreamsimlearningnewdimensions}
S.~Fu, N.~Tamir, S.~Sundaram, L.~Chai, R.~Zhang, T.~Dekel, and P.~Isola, ``Dreamsim: Learning new dimensions of human visual similarity using synthetic data,'' 2023.

\bibitem{perceptualLoss2016CVPR}
J.~Johnson, A.~Alahi, and L.~Fei-fei, ``Perceptual losses for real-time style transfer and super-resolution,'' in {\em CVPR}, 2016.

\bibitem{NIPS2012_c399862d}
A.~Krizhevsky, I.~Sutskever, and G.~E. Hinton, ``Imagenet classification with deep convolutional neural networks,'' in {\em Advances in Neural Information Processing Systems} (F.~Pereira, C.~Burges, L.~Bottou, and K.~Weinberger, eds.), vol.~25, Curran Associates, Inc., 2012.

\bibitem{dosovitskiy2021imageworth16x16words}
A.~Dosovitskiy, L.~Beyer, A.~Kolesnikov, D.~Weissenborn, X.~Zhai, T.~Unterthiner, M.~Dehghani, M.~Minderer, G.~Heigold, S.~Gelly, J.~Uszkoreit, and N.~Houlsby, ``An image is worth 16x16 words: Transformers for image recognition at scale,'' 2021.

\bibitem{liu2023grounding}
S.~Liu, Z.~Zeng, T.~Ren, F.~Li, H.~Zhang, J.~Yang, C.~Li, J.~Yang, H.~Su, J.~Zhu, {\em et~al.}, ``Grounding dino: Marrying dino with grounded pre-training for open-set object detection,'' in {\em ECCV}, 2024.

\bibitem{kirillov2023segany}
A.~Kirillov, E.~Mintun, N.~Ravi, H.~Mao, C.~Rolland, L.~Gustafson, T.~Xiao, S.~Whitehead, A.~C. Berg, W.-Y. Lo, P.~Doll{\'a}r, and R.~Girshick, ``Segment anything,'' 2023.

\end{thebibliography}

\end{document}


\title{MTReD: 3D Reconstruction Dataset for Fly-over Videos of Maritime Domain}
\author{Rui Yi Yong$^{1,2}$, Samuel Picosson$^1$, Arnold Wiliem$^{1,3}$
{\tt\small  }
\\ 
$^1$Shield AI \{rui.yi, samuel.picosson, arnold.wiliem\}@shield.ai \\
$^2$Monash University, Melbourne, Australia \\
$^3$Queensland University of Technology, Queensland, Australia
}  
\maketitle

These supplementary materials outline the details of the MTReD dataset. The video names and Youtube links of each video are provided in the firsrt two columns. Following this, the selected segments from each video are detailed as a set of ranges. For example [30, 190] would mean that we only extract images starting from the 30 second mark, up until the 190th second. Lastly, the frame skip frequency details how many frames are skipped before an image is saved, within each segment.

\begin{table}
  \centering
  {\small{
  \begin{tabular}{p{3cm}|p{7cm}|p{4.5cm}|p{2cm}}
     \toprule
     Video name & Youtube link & Selected segments (Seconds) & Frame Skip Frequency \\ \hline
    Vladi Strecker - Sailor's Dream & https://www.youtube.com/watch?v=QUrBPEa0R3Iz & [30, 190] & 2 \\ \hline
    Small Cargo Ship & https://www.youtube.com/watch?v=jhWQ8Ie2fE8 & [1, 14], [17, 50], [54, 95], [98, 105], [110, 115], [119, 137] & 2 \\ \hline
    Breathtaking Aerial Views of Cargo Ship Near Tacoma, WA & https://www.youtube.com/watch?v=JidbbnJXOB8 & [0, 127] & 2 \\ \hline
    Aerial view of Grounded Ship at theThousand Island Bridge & https://www.youtube.com/watch?v=oAKvdI34aNg & [0, 146] & 2 \\ \hline
    Drone footage of cargo ship at anchor  & https://www.youtube.com/watch?v=mQxYhry0ADE & [0, 177] & 4 \\ \hline
    A Bird's Eye View of the Beijing Spirit & https://www.youtube.com/watch?v=CZ-m3DknAeI & [0, 2], [4, 6], [14, 25], [34, 42], [44, 54], [55, 58], [67, 90], [95, 103], [170, 204] & 1 \\ \hline
    5 Cruise Ships docked in Nassau - 4K DRONE FOOTAGE & https://www.youtube.com/watch?v=3iqEq10G-Z8 & [2, 5], [8, 24], [27, 38], [40, 61], [64, 70], [73, 90], [95, 109], [112, 123], [127, 150], [155, 160], [163, 190], [193, 230], [232, 244], [248, 264], [270, 276] & 2 \\ \hline
    DJI MAVIC AIR - Small Island Drone Tour 4K - Izmir & https://www.youtube.com/watch?v=qMLdxcoRKp0 & [0, 57] & 1 \\ \hline
    Little Samphire Island Flyover, Fenit , County Kerry, Ireland - DJI Phantom 4 & https://www.youtube.com/watch?v=JLVYI8FwmlQ & [0, 2], [8, 123] & 1 \\ \hline
    Guamus - Cocos Island | Secret of South Pacific Place | Most Beautiful Ocean Scenery | 4K Drone & https://www.youtube.com/watch?v=NQbZTVFQda0 & [30, 120], [184, 484] & 4 \\ \hline
    Birds eye view of Hummocky Island & https://www.youtube.com/watch?v=giU0gnNLPpY & [0, 294] & 2 \\ \hline
    A Birds Eye View of Ram Island & https://www.youtube.com/watch?v=DsPAXaS8klE & [24, 180] & 1 \\ \hline
    Little Island (Pier 55) NYC Drone & https://www.youtube.com/watch?v=RbWChxZmTFU & [2, 230] & 2 \\ \hline
    FLY OVER BONDI BEACH - DJI MAVIC PRO & https://www.youtube.com/watch?v=WzdkbzMDEMY & [10, 170] & 2 \\ \hline
    6 July 24 San Pedro Drone. Lighthouse? & https://www.youtube.com/watch?v=fe68mC2-euQ & [0, 327] & 2 \\ \hline
    South Africa coast, beach ocean view, coastal flyover footage, mountains, high quality (4K Ultra-HD) & https://www.youtube.com/watch?v=o42mcNjSyZc & [0, 12], [15, 23], [26, 36], [38, 67], [69, 88], [91, 101], [104, 118], [122, 138], [141, 200] & 2 \\ \hline
    Hokkaido Coastline flyover sample & https://www.youtube.com/watch?v=5GlLVrRnJrw & [2, 57], [61, 85], [88, 194] & 2 \\ \hline
    Drone footage of the Perth coastline & https://www.youtube.com/watch?v=HMy52R2yLOA & [2, 7], [9, 12], [15, 18], [21, 25], [28, 35], [38, 44], [48, 55], [58, 62], [65, 73], [75, 82], [84, 92], [94, 112] & 1 \\ \hline
    The Jurassic Coast - Drone Footage & https://www.youtube.com/watch?v=iTuZCiOLLtk & [13, 250] & 2 \\ \hline
  \end{tabular}
  }}
  \caption{Content ranges for each data point in the proposed \NewDataset w along with reduction ratios to keep within memory constraints.}
  \label{tab:example}
\end{table}